\documentclass[fullpaper,cameraready]{nldl}

\usepackage[utf8]{inputenc}
\usepackage{url}
\usepackage{graphicx}
\usepackage{authblk}
\usepackage{amsmath}
\usepackage{amsfonts} 
\usepackage{threeparttable}

\usepackage{algorithm}
\usepackage{algorithmic}

\usepackage[square,numbers]{natbib}
\usepackage{url}
\usepackage{hyperref}
\usepackage{glossaries}     
\usepackage{graphicx}
\usepackage{amsmath}
\usepackage{multirow}
\usepackage{makecell}
\usepackage{xr}

\title{Boundary Aware U-Net for Glacier Segmentation}
\author[1]{Bibek Aryal\thanks{Corresponding Author: baryal@miners.utep.edu}}
\author[2]{Katie E. Miles}
\author[1]{Sergio A. Vargas Zesati}
\author[1]{Olac Fuentes}
\affil[1]{The University of Texas at El Paso, Texas, USA}
\affil[2]{Aberystwyth University, Aberystwyth, UK}

\date{\vspace{-5ex}}

\begin{document}

\newacronym{icimod}{ICIMOD}{International Centre for Integrated Mountain Development}
\newacronym{rsi}{RSI}{Remote Sensing Indexes}
\newacronym{iou}{IoU}{Intersection over Union}
\newacronym{dem}{DEM}{Digital Elevation Model}
\newacronym{hkh}{HKH}{Hindu Kush Himalayas}
\newacronym{ndvi}{NDVI}{Normalized Difference Vegetation Index}
\newacronym{ndwi}{NDWI}{Normalized Difference Water Index}
\newacronym{ndsi}{NDSI}{Normalized Difference Snow Index}
\newacronym{crs}{CRS}{Coordinate Reference System}
\newacronym{ss}{SS}{Saliency Score}
\newacronym{sm}{SM}{Saliency Map}
\nldlmaketitle

\begin{abstract}
  Large-scale study of glaciers improves our understanding of global glacier change and is imperative for monitoring the ecological environment, preventing disasters, and studying the effects of global climate change. Glaciers in the Hindu Kush Himalaya (HKH) are particularly interesting as the HKH is one of the world's most sensitive regions for climate change. In this work, we: (1) propose a modified version of the U-Net for large-scale, spatially non-overlapping, clean glacial ice, and debris-covered glacial ice segmentation; (2) introduce a novel self-learning boundary-aware loss to improve debris-covered glacial ice segmentation performance; and (3) propose a feature-wise saliency score to understand the contribution of each feature in the multispectral Landsat 7 imagery for glacier mapping. Our results show that the debris-covered glacial ice segmentation model trained using self-learning boundary-aware loss outperformed the model trained using dice loss. Furthermore, we conclude that red, shortwave infrared, and near-infrared bands have the highest contribution toward debris-covered glacial ice segmentation from Landsat 7 images.
\end{abstract}
\section{Introduction}

\begin{figure}[!h]
  \centering
  \includegraphics[width=\linewidth]{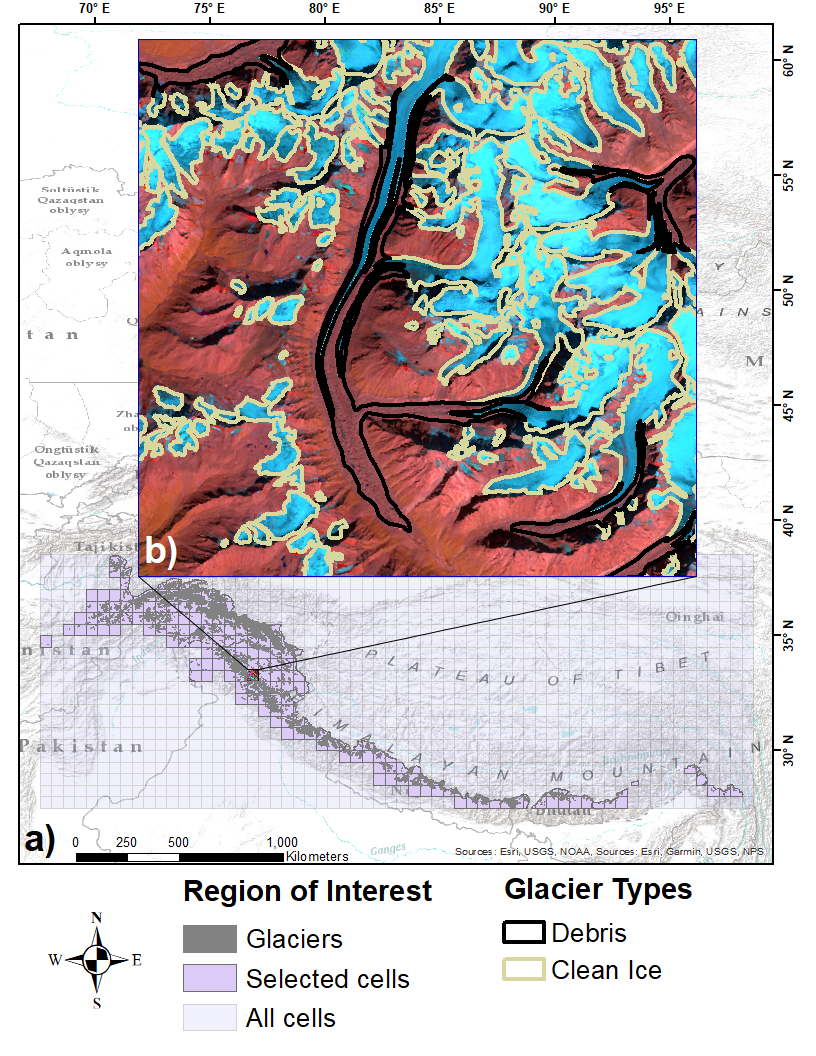}
  \caption{a) Spatially non-overlapping regions using fishnet grid. b) A zoomed image of one of the cells showing clean ice and debris glacier labels.}
  \label{fig:roi}
\end{figure}

Glacier delineation using remote sensing imagery has seen a growing use of deep learning in recent years~\cite{ baraka2020machine, florath2022glacier, he2020glacier, tian2022mapping, xie2020glaciernet}. This can be attributed to factors such as the availability of large-scale remote sensing data from multiple sources, the development of state-of-the-art deep learning architectures for image analysis, and the growing interest due to the impacts of climate change on glaciers in recent decades. The Himalaya is one of the world's most sensitive regions to global climate change, with impacts manifesting at particularly rapid rates~\cite{immerzeel2020importance, kraaijenbrink2017impact}. Unsurprisingly, much research has been focused on mapping glaciers in the \gls{hkh}~\cite{baraka2020machine, bhambri2009glacier, xie2020glaciernet}. Several studies have reported the performance of clean glacial ice and debris-covered glacial ice mapping in the \gls{hkh}; however, most research has been focused on specific glacier basins within the region and not across the region as a whole. 

Glaciers form when snow compresses under its own weight and hardens over long timescales~\cite{cuffey2010physics}. Near its formation, glacial ice has snow or ice surface cover and is known as clean glacial ice. As glacial ice slowly moves down  valleys under gravity, avalanches can deposit debris (rocks and sediment) on top of the glacier. Glacial ice having a significant covering of dirt/rocks/boulders  is known as debris-covered glacial ice. Clean glacial ice and debris-covered glacial ice appear differently in remote sensing imagery. The challenge lies in differentiating clean glacial ice from temporary snow/ice cover and debris-covered glacial ice from moraines and the surrounding valleys. The spectral uniqueness of clean glacial ice compared to surrounding terrain makes it relatively easy to identify and localize. However, the delineation of debris-covered glacial ice poses significant challenges because of its non-unique spectral signature. 

The earliest approaches for debris-covered glacial ice segmentation involving deep learning used multilayer perceptrons to estimate the supraglacial debris loads of Himalayan glaciers using pre-defined glacier outlines~\cite{bishop1995spot, bishop1999spot}. More recent methods for glacier segmentation use Convolutional Neural Networks (CNNs) due to their success in image-based applications~\cite{baraka2020machine, mohajerani2019detection, xie2020glaciernet, zheng2021interactive}. Most recent approaches to learning-based  image segmentation use variants of the U-Net architecture~\cite{ronneberger2015u}. Originally introduced for biomedical image segmentation, the U-Net has seen successes in numerous applications involving satellite image segmentation~\cite{ mcglinchy2019application, robinson2020human, zhang2018urban}. Moreover, different architectures based on the U-Net have also been used for glacier segmentation in recent years~\cite{baraka2020machine, xie2020glaciernet}. However, unlike segmentation for general images, the results when it comes to glacier segmentation are not very good, particularly for debris-covered glacial ice.

Here we present a variant of the U-Net and train it using multispectral images from Landsat 7 as inputs. Researchers have shown that the performance of deep learning models can be improved by learning multiple objectives from a shared representation~\cite{caruana1997multitask}. Early approaches to learn multiple tasks use weighted sum of losses, where the loss weights are either constant or manually tuned~\cite{eigen2015predicting, leibe2008robust, sermanet2013overfeat}. We propose a method to combine two different loss functions - masked dice loss~\cite{aryal2021semi} and boundary loss~\cite{bokhovkin2019boundary} - to simultaneously learn multiple objectives automatically during the training process for improved performance.

While deep learning models have been shown to perform well on various tasks involving computer vision, the interpretability of these models is limited. Deep neural networks are often considered black boxes, since their decision rules can not be described easily. Unlike coefficients and decision boundaries of simpler machine learning methods such as linear regression and decision trees, weights of neurons in deep neural networks can not be understood as knowledge directly. The development of transparent, understandable, and explainable models is imperative for the wide-scale adoption of deep learning models. Over the years, many have proposed different approaches to describe deep leaning models~\cite{luo2016understanding, yosinski2015understanding, zheng2021interactive}. One of the most widely used methods to envision which pixels in the input image affect the outputs the most is by visualizing saliency maps~\cite{simonyan2013deep}. A saliency map is obtained by calculating the gradient of the given output class with respect to the input image by letting gradients backpropagate to the input. In the case of multispectral or hyperspectral images, spectral saliency~\cite{6515145} is used to visualize salient pixels of an image. Image saliency maps, computed independently for all channels on a multispectral image, can be used to visualize the contribution of each pixel in each channel toward the final output. We propose a method to quantify each channel's contributions towards the final label in the context of glacier segmentation using Landsat 7 imagery. 

\section{Dataset and Methodology}

The \gls{hkh} region covers an area of about 4.2 million $km^2$ from about $15^{\circ}$ to $39^{\circ}$ N latitude and about $60^{\circ}$ to $105^{\circ}$ E longitude extending across eight countries consisting of Afghanistan, Bangladesh, Bhutan, China, India, Myanmar, Nepal, and Pakistan~\cite{bajracharya2011status}. The geographic extent of the glaciers within the \gls{hkh}, however, ranges from about $27^{\circ}$ to $38^{\circ}$ N and about $67^{\circ}$ to $98^{\circ}$ E (Figure~\ref{fig:roi}).

We downloaded the Landsat 7 images used for label creation using Google Earth Engine. Landsat 7 contains the Enhanced Thematic Mapper Plus (ETM+) sensor which captures multiple spectral bands as shown in Table~\ref{tab:landsat7_bands}. The thermal infrared bands were upsampled from 60 meters to 30 meters resulting in all bands having a spatial resolution of 30 meters. The glacier outlines (labels)~\cite{bajracharya2011clean} were downloaded from \gls{icimod} Regional Database System. (\url{http://rds.icimod.org/Home/DataDetail?metadataId=31029}) The glacier labels contain information on clean-ice and debris-covered glaciers in the \gls{hkh} for regions within Afghanistan, Bhutan, India, Nepal, and Pakistan. The \gls{icimod} glacier outline labels used in this research were derived using the object-based image classification methods separately for clean-ice and debris-covered glaciers and fine-tuned with manual intervention~\cite{bajracharya2011status}.

\begin{table}[!h]
  \centering
  \caption{Landsat 7 bands description}
  \label{tab:landsat7_bands}
  \begin{tabular}{|l|l|l|} \hline
    Name & Description \\ \hline
    B1 & Blue \\
    B2 & Green \\
    B3 & Red \\
    B4 & Near Infrared \\
    B5 & Shortwave Infrared 1 \\
    B6\_VCID\_1 & Low-gain Thermal Infrared \\
    B6\_VCID\_2 & High-gain Thermal Infrared \\
    B7 & Shortwave Infrared 2 \\
    \hline
  \end{tabular}
\end{table}

\begin{figure}[!h]
  \centering
  \includegraphics[width=\linewidth]{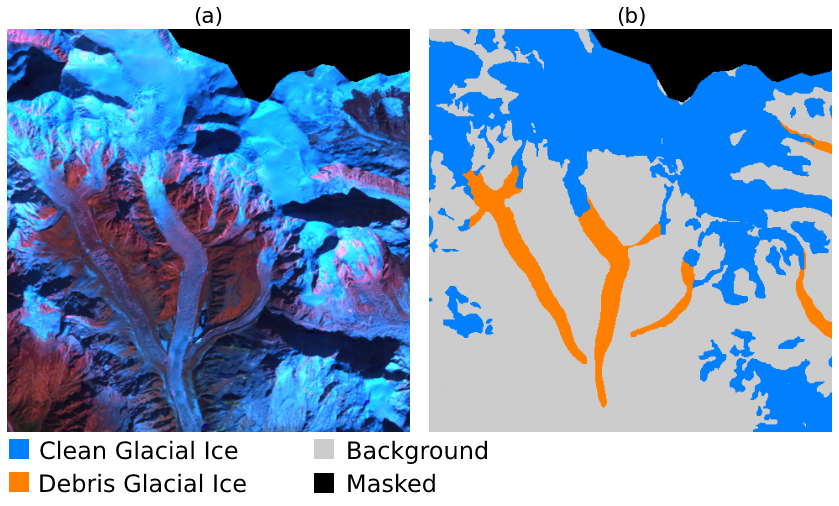}
  \caption{(a) Sample sub-image, (b) Corresponding Clean Glacial Ice, Debris Glacial Ice, Background, and Masked labels}
  \label{fig:label_classes}
\end{figure}

The Landsat 7 images that were used for delineating glacier labels~\cite{bajracharya2011status} overlap spatially. To avoid spatial overlap between train and test regions, we created polygon features representing a fishnet of rectangular cells for the entire geographical region. We then created a mosaic of all Landsat 7 images used for labeling into a single raster and clipped the raster mosaic to country boundaries for glacier labels (Figure~\ref{fig:roi}) to avoid false negative glacier labels in the dataset. Finally, we discarded the rasters within the polygon cells that do not contain any glacier labels and downloaded clipped regions within selected cells. The Google Earth Engine code to replicate this process can be found in  repository \url{https://code.earthengine.google.com/?accept_repo=users/bibekaryal7/get_hkh_tiff}. The selected cells were then randomly split into train, validation, and test sets with no geospatial overlap. 1163 out of 1364 cells were filtered out to leave us with 141, 20, and 40 cells in the training, validation, and test sets respectively. Each cell was then cropped into multiple sub-images of $512\times512$ pixels and the sub-images with less than $10\%$ of pixels as glacier labels were discarded to reduce class imbalance. These sub-images are then normalized and provided as input to the model. There are 333, 68, and 98 sub-images in the training, validation, and test sets respectively. Every pixel within each sub-image can have one of four different classes as can be seen in Figure~\ref{fig:label_classes}.

\begin{figure}[!h]
  \centering
  \includegraphics[width=\linewidth]{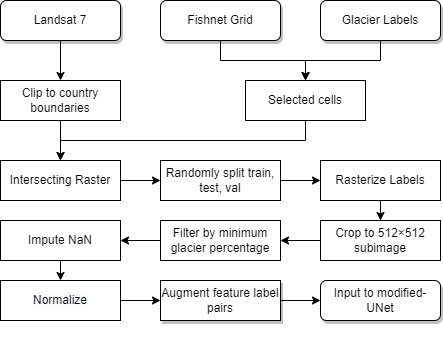}
  \caption{Input preprocessing}
  \label{fig:preprocessing}
\end{figure}

The step-by-step processing we followed to prepare input features for the model is shown in Figure~\ref{fig:preprocessing}. The distribution of pixels for train, validation, and test set across different classes is shown in Table~\ref{tab:label-distribution} and highlights that the distribution of pixels across different sets is similar and labels are heavily imbalanced across classes.

\begin{table}[!h]
  \centering
  \caption{Labels Distribution - Random Sampling}
  \label{tab:label-distribution}
  \begin{threeparttable}
  \begin{tabular}{|l|l|l|l|l|} \hline
    split & background & clean & debris & masked \\ \hline
    train & 72.44\% & 21.77\% & 2.44\% & 3.35\% \\
    val & 68.69\% & 23.22\% & 3.24\% & 4.85\% \\
    test & 70.16\% & 22.97\% & 2.65\% & 4.21\% \\
    \hline
  \end{tabular}
  \begin{tablenotes}[para,flushright]
     \raggedright clean = clean glacial ice \\
     debris = debris-covered glacial ice 
  \end{tablenotes}
  \end{threeparttable}
\end{table}

\begin{figure}[!h]
  \centering
  \includegraphics[width=\linewidth]{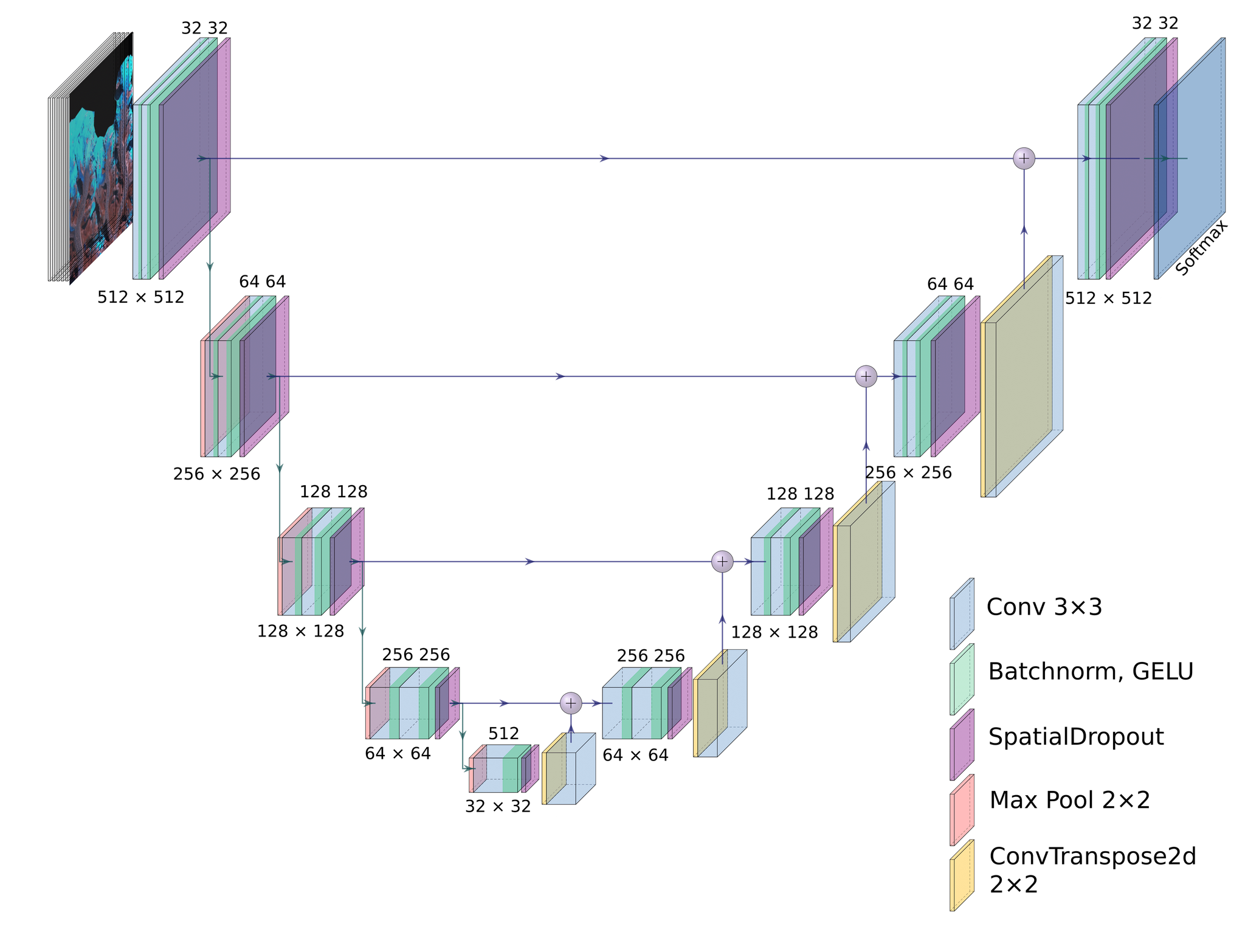}
  \caption{Our modified U-Net architecture has 32 feature maps in the first convolution layer. We also introduce batch normalization and spatial dropout in the modified architecture.}
  \label{fig:architecture}
\end{figure}

We used a modified version of the U-Net architecture~\cite{ronneberger2015u} as shown in Figure~\ref{fig:architecture}. Each input sub-image is $512\times512$ pixels in size. Zero padding was added during each convolution operation to make the output labels the same size as input sub-images. We replaced the Rectified Linear Unit (ReLU) in the original U-Net architecture with Gaussian Error Linear Units (GELU)~\cite{hendrycks2016gaussian}. We applied batch normalization after each convolution operation and spatial dropout~\cite{tompson2015efficient} of $0.1$ after  each down-sampling and up-sampling block to reduce overfitting. We also randomly modified $15\%$ of the training samples by either rotating $(90^{\circ}, 180^{\circ}, 270^{\circ})$ or flipping (horizontal/vertical) the input sub-images to the model. We trained the modified U-Net architecture for $250$ epochs using the Adam optimizer and evaluated the performance based on precision, recall, and \gls{iou}.

We trained two separate models, one for segmenting clean glacial ice and one for debris-covered glacial ice,  and combined the outputs to produce the final segmentation map. Definitions of what constitutes debris-covered glacial ice vary widely, however, as a glacier does not have to be fully covered by debris to be classified as debris-covered glacial ice~\cite{MILES2020103212}. Therefore, for the pixels where debris-covered glacial ice labels overlapped with clean glacial ice labels on the final segmentation map, the output label was set as debris-covered glacial ice. The code to
replicate our process can be found in the GitHub repository (\url{https://github.com/Aryal007/glacier_mapping.git}).
\section{Experiments}

\subsection{Self-learning Boundary-aware Loss}

The subject of this section of our work lies at the intersection of two branches of research, which are penalizing misalignment of label boundaries by using a boundary-aware loss and learning multi-task weights during the training process. We propose a combined loss ($\mathcal{L}_{Combined}$) that is a weighted sum of masked dice loss $(\mathcal{L}_{MDice})$ and boundary loss $(\mathcal{L}_{Boundary})$, as described in Equation 1. We also compare the performance of our methods using the modified U-Net to the standard U-Net trained on cross entropy loss $\mathcal({L}_{CE})$.

\begin{equation}
  \begin{aligned}
    & \mathcal{L}_{Combined} = \alpha\times\mathcal{L}_{MDice} + (1 - \alpha)\times\mathcal{L}_{Boundary} &
  \end{aligned}
\end{equation}

\begin{table*}[ht]
\centering
\caption{Performance comparisons between standard U-Net trained using cross entropy loss and modified U-Net trained using combined loss and self-learning boundary-aware loss}
\label{tab:boundaryloss}
\begin{threeparttable}
\begin{tabular}{|l|l|l|l|l|l|l|l|} \hline
\multirow{2}{*}{Loss ($\mathcal{L}$)} & \multirow{2}{*}{$\mathcal{L}_{weight(s)}$} & \multicolumn{3}{c|}{clean}  & \multicolumn{3}{c|}{debris}   \\  \cline{3-8}
& & \multicolumn{1}{l|}{Precision} & \multicolumn{1}{l|}{Recall} & \gls{iou} & \multicolumn{1}{l|}{Precision} & \multicolumn{1}{l|}{Recall} & \gls{iou} \\ \hline
$\mathcal{L}_{CE}$ & $-$ & \multicolumn{1}{l|}{89.39\%} & \multicolumn{1}{l|}{84.65\%} & 70.16\%  & \multicolumn{1}{l|}{0.00\%} & \multicolumn{1}{l|}{0.00\%} & 0.00\% \\ 
$\mathcal{L}_{Combined}$ & $0$ & \multicolumn{1}{l|}{68.40\%} & \multicolumn{1}{l|}{0.25\%} & 0.25\%  & \multicolumn{1}{l|}{3.00\%} & \multicolumn{1}{l|}{99.75\%} & 3.00\% \\ 
$\mathcal{L}_{Combined}$ & $0.1$ & \multicolumn{1}{l|}{79.82\%} & \multicolumn{1}{l|}{79.66\%} & 66.31\%  & \multicolumn{1}{l|}{50.93\%} & \multicolumn{1}{l|}{49.30\%} & 33.43\% \\ 
$\mathcal{L}_{Combined}$ & $0.5$ & \multicolumn{1}{l|}{81.60\%} & \multicolumn{1}{l|}{80.77\%} & 68.33\%  & \multicolumn{1}{l|}{51.16\%} & \multicolumn{1}{l|}{46.92\%} & 32.41\% \\ 
$\mathcal{L}_{Combined}$ & $0.9$ & \multicolumn{1}{l|}{81.60\%} & \multicolumn{1}{l|}{80.77\%} & 68.33\%  & \multicolumn{1}{l|}{51.16\%} & \multicolumn{1}{l|}{46.92\%} & 32.41\% \\ 
$\mathcal{L}_{Combined}$ & $1$ & \multicolumn{1}{l|}{80.31\%} & \multicolumn{1}{l|}{80.65\%} & 67.34\%  & \multicolumn{1}{l|}{46.00\%} & \multicolumn{1}{l|}{44.10\%} & 29.05\% \\ 
$\mathcal{L}_{SLBA}$ & $Dynamic$ & \multicolumn{1}{l|}{81.59\%} & \multicolumn{1}{l|}{80.55\%} & 68.17\%  & \multicolumn{1}{l|}{51.97\%} & \multicolumn{1}{l|}{53.81\%} & 35.94\% \\ \hline
\end{tabular}
\begin{tablenotes}[para,flushright]
 \raggedright *clean = clean glacial ice; 
 debris = debris-covered glacial ice 
\end{tablenotes}
\end{threeparttable}
\end{table*}

The value of hyperparameter $\alpha$ can be set manually between 0 and 1. Having an $\alpha$ of 1 is equivalent to training the model exclusively using masked dice loss and an $\alpha$ of 0 is the same as training the model exclusively using boundary loss. However, tuning the value of $\alpha$ manually for the best results is an expensive process. In order to learn the weights for $\mathcal{L}_{Boundary}$ and $\mathcal{L}_{MDice}$ through backpropagation, we initially set $\alpha$ to 0.5 and let the model find the best value of $\alpha$. However, we observe that without any constraints on the value of $\alpha$, the network updates $\alpha$ such that $\mathcal{L}_{Combined}$ is minimized without necessarily having to minimize $\mathcal{L}_{MDice}$ or $\mathcal{L}_{Boundary}$. This results in poor performance. Inspired by \cite{Kendall_2018_CVPR} for weighing two different loss functions, we propose Self-Learning Boundary-Aware loss ($\mathcal{L}_{SLBA}$) that is a combination of $\mathcal{L}_{MDice}$ and $\mathcal{L}_{Boundary}$. 

\begin{equation}
  \label{eqn:self_aware_boundaryawareloss}
  \begin{aligned}
    & \resizebox{0.5\textwidth}{!}{$\mathcal{L}_{SLBA} = \frac{1}{2\alpha_1^2}\times\mathcal{L}_{MDice} + \frac{1}{2\alpha_2^2}\times\mathcal{L}_{Boundary} + \lvert \ln{(\alpha_1\times\alpha_2)}\rvert$}&
  \end{aligned}
\end{equation}

In the case of $\mathcal{L}_{SLBA}$, $\alpha_1$ and $\alpha_2$ both are initially set to $1$ and we let the model find the best value for $\alpha_1$ and $\alpha_2$ through backpropagation. In Table~\ref{tab:boundaryloss} we show performance for different values of $\alpha$ in the case of $\mathcal{L}_{Combined}$ and performance of $\mathcal{L}_{SLBA}$. One advantage of using $\mathcal{L}_{SLBA}$ over $\mathcal{L}_{Combined}$ is that there is no extra hyperparameter that requires fine-tuning. All experiments in Table~\ref{tab:boundaryloss} use eight features from Landsat 7 imagery as inputs.

\begin{figure}[!h]
  \centering
  \includegraphics[width=\linewidth]{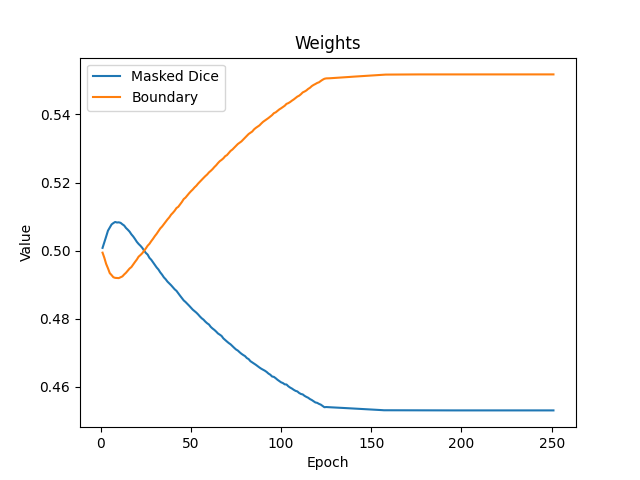}
  \caption{Masked Dice Loss weights and Boundary Loss weights vs.\@ epoch for debris-covered glacial ice}
  \label{fig:sigma}
\end{figure}

From Table~\ref{tab:boundaryloss}, we see that $\mathcal{L}_{SLBA}$ performs the best for debris-covered glacial ice segmentation and eliminates the need to fine-tune loss weights. We can also see that the model fails to converge when training solely on boundary loss ($\alpha = 0$) and training on glacier boundaries by incorporating boundary loss along with masked dice loss results in an overall improvement in performance for debris-covered glacial ice regardless of the weighting factor. Figure~\ref{fig:sigma} shows the weights for masked dice loss ($\frac{1}{2\alpha_1^2}$) and the weights for boundary loss ($\frac{1}{2\alpha_2^2}$) vs. epoch during training for debris-covered glacial ice. The optimal values for $\alpha_1$ and $\alpha_2$ are calculated to be $0.9569$ and $1.045$ for clean glacial ice segmentation and $0.952$ and $1.05$ for debris-covered glacial ice segmentation for $\mathcal{L}_{SLBA}$.

\subsection{Representation Analysis}

To understand the contribution of each feature in the multispectral image toward the final label, we computed a \gls{ss} for each feature by summing all pixels in the \gls{sm} for that feature.

\begin{figure}[!h]
  \centering
  \includegraphics[width=\linewidth]{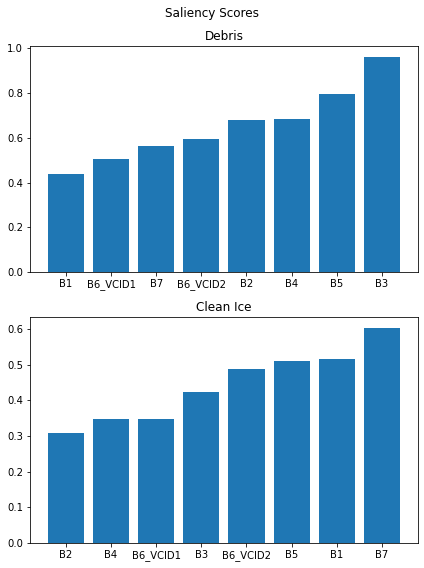}
  \caption{Average saliency scores for all sub-images in training set.}
  \label{fig:saliency_scores}
\end{figure}

\begin{equation}
  \begin{aligned}
    & \resizebox{0.5\textwidth}{!}{$\gls{ss}_{feature} = \sum\limits_{\substack{i=0}}^{\substack{c-1}}\sum\limits_{\substack{j=0}}^{\substack{r-1}} \gls{sm}_{feature}(i, j) \forall~ feature \in Input$}&
  \end{aligned}
\end{equation}
where: 
\begin{tabbing}
$r, c$ = number of rows, columns in saliency map\\
\end{tabbing}

Average feature saliency scores across all the images in the training samples are shown in Figure~\ref{fig:saliency_scores}. The channel-wise contributions towards debris-covered glacial ice segmentation in decreasing order are: red, shortwave infrared 1, near infrared, green, high-gain thermal infrared, shortwave infrared 2, low-gain thermal infrared, and blue. Similarly, for clean glacial ice segmentation, the channel-wise contributions in decreasing order are: shortwave infrared 2, blue, shortwave infrared 1, high-gain thermal infrared, red, low-gain thermal infrared, near infrared, and green. As shown in Figure~\ref{fig:saliency_scores}, the segmentation models have different high contributing channels for clean glacial ice and debris-covered glacial ice segmentation.
\section{Discussion}

Glaciers have been melting at an unprecedented rate in recent years due to global climate change~\cite{immerzeel2020importance, kraaijenbrink2017impact}. Glaciers are the largest freshwater reservoir on the planet~\cite{shiklomanov1990global}, so it is necessary to understand the changes they undergo. As a result, numerous approaches to automatically delineate glacier boundaries have been proposed~\cite{baraka2020machine, bhambri2009glacier, florath2022glacier,  he2020glacier,  tian2022mapping, xie2020glaciernet}. We frequently observe deep learning methods outperforming traditional machine learning methods for glacier segmentation in the literature~\cite{aryal2020glacier, baraka2020machine}. However, the results have not been very good, particularly in the case of debris-covered glacial ice. 

In this work, we modify U-Net and train it using a novel loss function that allows the modified U-Net to focus on glacier boundaries. From Table~\ref{tab:boundaryloss}, we see that standard U-Net is not able to detect debris-covered glacial ice in input sub-images. We can also see from Table~\ref{tab:label-distribution} that only 2.44\% of pixels in the training set correspond to debris-covered glacial ice. This shows that our proposed method is more robust than the original U-Net to imbalanced labels, which are common in  remote sensing datasets.

From Figure~\ref{fig:sigma}, we can see how the weights change for $\mathcal{L}_{SLBA}$ while training. A higher weight is assigned to masked dice loss at the beginning and the weights for boundary loss are gradually increased during training. The reason behind this could be that for an untrained model, it may be easier to learn glacier instances over trying to learn the boundaries. However once the network learns to label instances, it is easier to learn the glacier boundaries. This also explains why the model fails to converge when training solely on $\mathcal{L}_{Boundary}$ from scratch as can be seen from the results in Table~\ref{tab:boundaryloss}.

We presented methods to improve debris-covered glacial ice segmentation from remote sensing imagery using deep learning. While we were able to show significant improvements over existing methods, the \gls{iou} for debris-covered glacial ice still leaves much to be desired. The existing body of literature on the topic has shown that the performance for debris-covered glacial ice segmentation can be improved by incorporating thermal signatures~\cite{ranzi2004use} and topographical information~\cite{bolch2005glacier, molg2018consistent, paul2004combining} from other satellites. Since debris-covered glacial ice is common in low-gradient areas due to how it forms and has cooler surface temperatures compared to the surrounding non-glaciated regions, we suspect that adding this information can further help improve the performance of debris-covered glacial ice segmentation. We may also be able to see an improvement in performance by using images from the recently-launched Landsat 9 satellite, instead of the Landsat 7 images used in this work. The Operational Land Imager 2 (OLI-2) and the Thermal Infrared Sensor 2 (TIRS-2) sensors on Landsat 9 provide data that is radiometrically and geometrically superior to instruments on the previous generation Landsat satellites. With the higher radiometric resolution, Landsat 9 can differentiate 16,384 shades of a given wavelength compared to only 256 shades in Landsat 7. Meanwhile, the TIRS-2 in Landsat 9 enables improved atmospheric correction and more accurate surface temperature measurements. Future work includes using the images captured through these improved sensors and incorporating additional information such as a digital elevation model for improving debris-covered glacial ice segmentation performance.
\section{Conclusion}

\begin{figure}[!h]
  \centering
  \includegraphics[width=\linewidth]{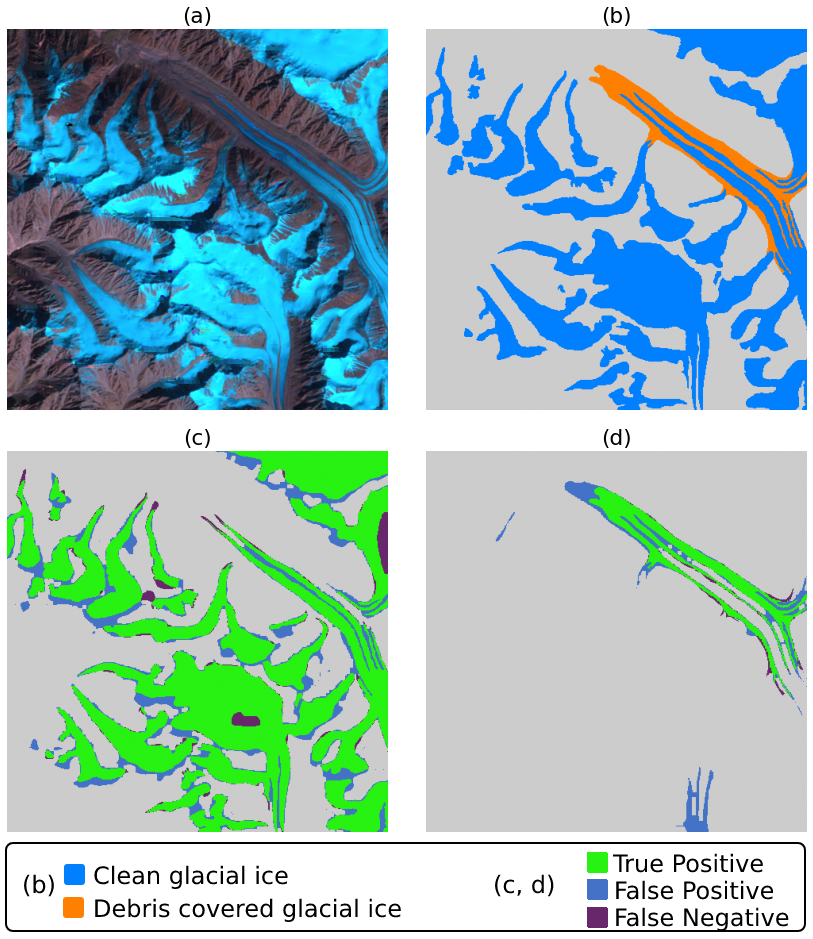}
  \caption{(a) Sample subimage from test set (b) Corresponding clean glacial ice and debris-covered glacial ice ground truth labels (c) True positive (TP), False positive (FP), False negative (FN) for clean glacial ice (\gls{iou} $79.17\%$) (d) TP, TP, FN for debris-covered glacial ice (\gls{iou} $59.19\%$)}
  \label{fig:sample_pred}
\end{figure}

In this research study, we proposed a modified version of the U-Net architecture for large-scale debris-covered glacial ice and clean glacial ice segmentation in the \gls{hkh} from Landsat 7 multispectral imagery and concluded that debris-covered glacial ice (\gls{iou}: $35.94\%$) is significantly harder to delineate compared to clean glacial ice (\gls{iou}: $68.17\%$)(Table~\ref{tab:boundaryloss}). We also introduced two different methods to combine commonly-used masked dice loss and boundary loss to incorporate label boundaries into the training process. We show that the performance of debris-covered glacial ice segmentation can be improved by encouraging the deep learning model to focus on label boundaries. The performance can be improved further by correctly weighing loss terms. Furthermore, the relative weights can be learned automatically from the data during the training process using our proposed loss ($\mathcal{L}_{SLBA}$). Figure~\ref{fig:sample_pred} shows the performance of the models trained using $\mathcal{L}_{SLBA}$ on a sample image from the test set. We also introduced the concept of feature saliency scores to quantify the contribution of each feature (channel) in the input image toward the final label and concluded that the red, shortwave infrared, and near infrared bands contribute the most towards the final label for debris-covered glacial ice segmentation, while shortwave infrared 2, blue, shortwave infrared 1 bands contributed the most towards the final label for clean glacial ice segmentation.

\section{Acknowledgements}

We would like to thank Microsoft for providing us with the Microsoft Azure resources through their AI for Earth grant program (Grant ID: AI4E-1792-M6P7-20121005). We also acknowledge \gls{icimod} for providing a rich dataset which this work has been built on. This research was supported in part by the Department of Computer Science at The University of Texas at El Paso.
\bibliographystyle{abbrvnat}
\bibliography{references}

\end{document}